\documentclass{article}
\usepackage[T1]{fontenc}
\usepackage[utf8]{inputenc}
\usepackage[]{ismir} 
\usepackage{amsmath,cite,url,multirow,amsfonts}
\usepackage{graphicx}
\usepackage{color}

\usepackage{makecell, booktabs}

\usepackage{microtype}

\title{Optical Music Recognition of Jazz Lead Sheets}

\multauthor
  {Juan C. Martinez-Sevilla$^1$ \hspace{1cm} Francesco Foscarin$^{2}$ \hspace{1cm} Patricia Garcia-Iasci$^{1, 3}$}
  {{\bf David Rizo$^{1, 4}$ \hspace{1cm} Jorge Calvo-Zaragoza$^1$ \hspace{1cm} Gerhard Widmer$^2$}\\
  $^1$ Pattern Recognition and Artificial Intelligence Group, University of Alicante, Spain\\
  $^2$ Institute of
Computational Perception, Johannes Kepler University Linz, Austria\\
  $^3$ University of Salamanca, Spain\\
  $^4$ Instituto Superior de Enseñanzas Artísticas de la Comunidad Valenciana, EASDA, Spain\\
  {\tt\small jcmartinez.sevilla@ua.es}
}

\def\authorname{J. C. Martinez-Sevilla, F. Foscarin, P. Garcia-Iasci, D. Rizo, J. Calvo-Zaragoza, and G. Widmer}

\usepackage[bookmarks=false,pdfauthor={\authorname},pdfsubject={\pdfsubject},hidelinks]{hyperref}

\begin{document}

\maketitle

\begin{abstract}

%
In this paper, we address the challenge of Optical Music Recognition (OMR) for handwritten jazz lead sheets, a widely used musical score type that encodes melody and chords. The task is challenging due to the presence of chords, a score component not handled by existing OMR systems, and the high variability and quality issues associated with handwritten images. Our contribution is two-fold. We present a novel dataset consisting of 293 handwritten jazz lead sheets of 163 unique pieces, amounting to 2021 total staves aligned with Humdrum **kern and MusicXML ground truth scores. We also supply synthetic score images generated from the ground truth. The second contribution is the development of an OMR model for jazz lead sheets. We discuss specific tokenisation choices related to our kind of data, and the advantages of using synthetic scores and pretrained models. We publicly release all code, data, and models.\footnote{\url{https://grfia.dlsi.ua.es/jazz-omr/}}

\end{abstract}

\section{Introduction}\label{sec:introduction}

A lead sheet is a kind of sheet music (musical score) that encodes the melody, chords, and sometimes lyrics of a music composition. Opposed to music styles, such as classical music, where the composer specifies with a high degree of precision what musicians have to play, lead sheets are popular in contexts where a lot of freedom is given to the performer. This is the case of jazz music, which lists improvisation as one of its core elements.
Multiple collections of ``unofficial'' lead sheets (i.e., not released by the original authors), named \textit{Fake Books}, were created by manual transcription from the performed tracks and are widely employed by jazz musicians during learning, performing, and composing~\cite{FosterD21}.

\begin{figure}
  \centering
  \includegraphics[alt={Prelude To a Kiss Handwritten Lead Sheet},width=1\linewidth]{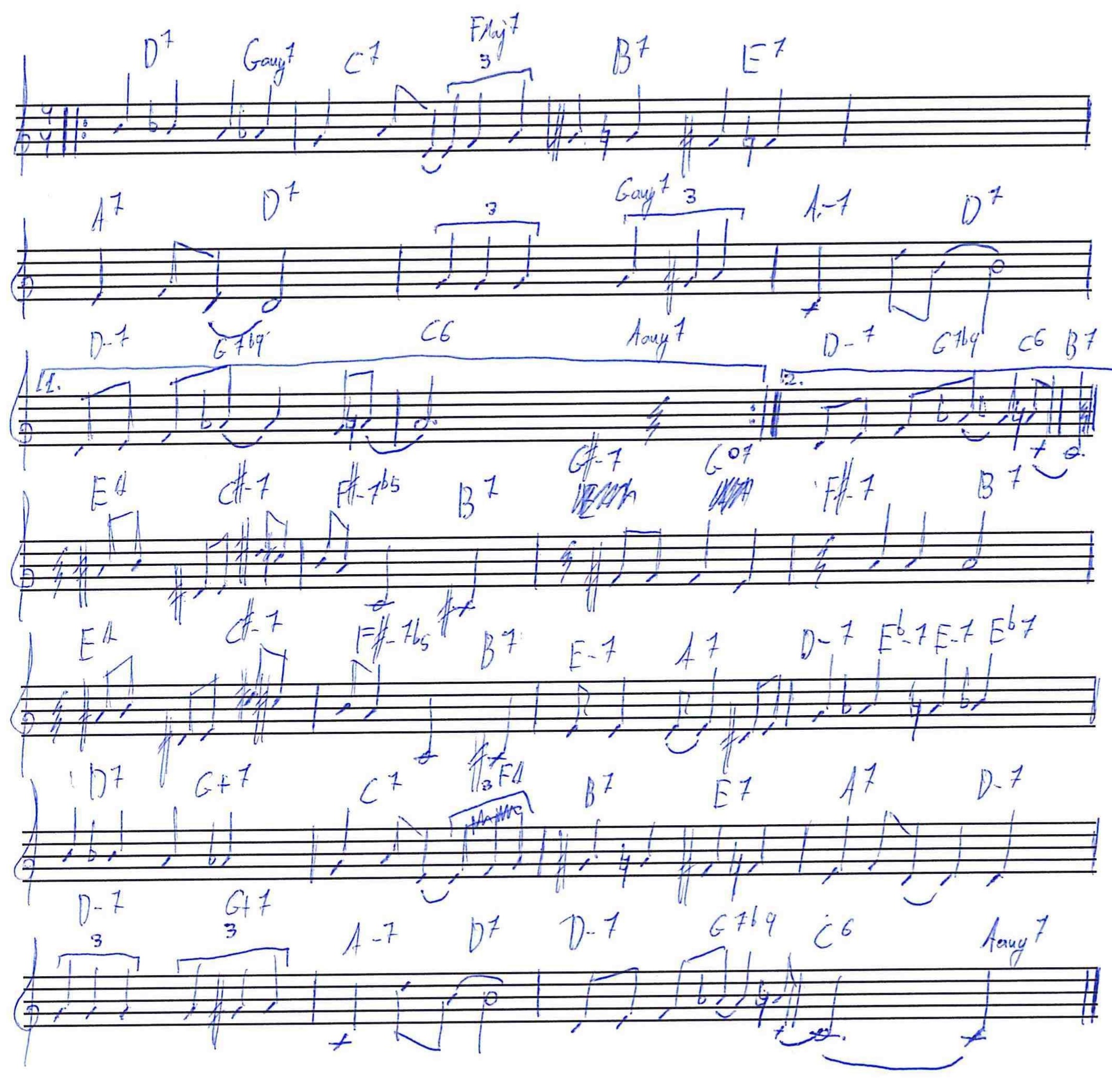}
  \caption{A (particularly problematic) handwritten lead sheet of Duke Ellington's ballad \textit{Prelude to a Kiss}.}
  \label{fig:prelude_sheet}
\end{figure}

However, handwritten or printed lead sheets still have limited utility compared to \textit{digitised} lead sheets, which store musical elements in a format that is easily accessible for both people and computers. For example, a digitised lead sheet enables operations such as corrections, transposition, sonification of melody and chords, changing the layout, extension into a longer arrangement, and use in automatic systems for analysis, retrieval, and accompaniment \cite{Calvo-ZaragozaH20understanding}.

Producing a digitised lead sheet is possible with music notation software such as MuseScore, Finale, Dorico, etc. However, musicians often find it faster or more enjoyable to write a lead sheet on paper by hand. 
This is where Optical Music Recognition (OMR) technology becomes useful in the context of jazz lead sheets, as it promises to convert handwritten ones into a digitised format.

However, this task presents unique challenges. The first is the variability in handwriting styles for all the musical symbols, paired with the necessity of handling ``dirty'' notation such as crossing out and corrections (see \figref{fig:prelude_sheet}). The second is the handling of chord symbols, which, to the best of our knowledge, are not included in any existing OMR system. This includes correctly predicting a musically valid chord name, but also correctly aligning the chord to its position in the musical score. Using heuristics based on the vertical alignment is often not sufficient, as we can see in the score of~\figref{fig:prelude_sheet} where the first two chords are almost on top of the 2nd and 4th note, even if in the ground truth, they should be aligned to the 1st and 3rd note. Moreover, we need to make some original considerations on the tokenisation process for chords and melodies, i.e., on how to effectively preprocess them for the usage in our network. Finally, there isn't a currently available dataset of handwritten jazz lead sheets and digitised ground truth specifically devised for OMR tasks. 

In this paper, we focus on lead sheets with only melody and chords (leaving lyrics for a future extension) and address the aforementioned challenges. We present a new dataset of 293 handwritten lead sheets, collected from jazz school students and professional musicians from Spanish institutions, which include melody and chords (see Section~\ref{sec:dataset}). We propose an end-to-end
OMR model (see Section~\ref{sec:methodology}) dedicated to these data, and propose a set of experiments (see Section~\ref{sec:experiments}) to analyse the impact of different tokenisation strategies, the importance of pretraining, and the usage of synthetic data.

\section{Related Work}
Our problem of transcribing melody and aligned lyrics is very similar to the transcription of melody and lyrics targeted by Martinez-Sevilla et al.~\cite{martinez2023}. They develop a network that, given a music region rotated by 90 degrees it performs an internal reshaping of the hidden space, to horizontally slice the score at note position. The result is a sequential hidden space that alternatively encodes notes and lyrics. A similar approach was proposed for polyphonic music by R{\'\i}os-Vila et al.~\cite{rios2023}. Although our lead sheets, with chords positioned above the notes, may appear similar and could suggest a comparable approach, there is a fundamental distinction. The earlier studies relied on synthetic data with perfect vertical alignment, while our handwritten data exhibits vertical misalignment between chords and notes, rendering this approach unsuitable.

A more promising system, which is currently state-of-the-art for full-page OMR for polyphonic music, is the Sheet Music Transformer~\cite{Rios24}, where the geometric considerations described above are left out in favor of an autoregressive transformer, which cross-attends to the encoded input image and learns to directly produce an output in Humdrum **kern format \cite{Huron1997}, a compact encoding for musical scores. While we are not targeting polyphonic or full-page recognition, we reuse their network architecture, since it is flexible enough to be adapted to our use case, and it has the potential to be easily scaled for future systems.

Regarding our proposed dataset, there are no other publicly available collections of handwritten lead sheets. A related work is the CoCoPops dataset~\cite{arthur2023cocopops}, a recent collection of digitised lead sheets. While there are no overlaps between their and our pieces, we support their effort of uniforming the encoding and we make available our digitised scores in the same format, i.e., Humdrum **kern files, with Harte~\cite{harte2005symbolic} syntax for chords. 
The Harte syntax is also used by the ChoCo dataset~\cite{deberardinis2023choco}, which also partially overlaps with our dataset, though their interest is mainly on chords.

\section{Dataset}\label{sec:dataset}

The dataset we release consists of images of musical scores, aligned with a digitised version. Every score encodes a lead sheet of a jazz standard, consisting of a monophonic melody\footnote{Few samples have polyphonic melodies.} and chord symbols.

\subsection{Digitised musical scores}

We provide musical scores for 163 unique jazz standards in MusicXML and Humdrum **kern format. The latter is widely used in systems that output musical scores \cite{Rios24}, because it is a compact and easy-to-handle format.
The MusicXML scores are taken from the Wikifonia database (discontinued in 2013) and partially corrected. We also leave the lyrics, if present in the original files, as they could be helpful for future extensions, but we don't consider them in this work.

We convert the MusicXML scores to **kern with the \emph{musicxml2hum} tool in Humlib. \footnote{https://github.com/craigsapp/humlib (Retrieved September 10, 2024)} However, the original handling of chords creates an \texttt{**mxhm} spine with root, type, and bass, but completely discards the chord extensions (which are an important part of jazz chords). Moreover, it translates the MusicXML  \texttt{harmony/kind}\footnote{\href{https://www.w3.org/2021/06/musicxml40/musicxml-reference/elements/harmony}{https://www.w3.org/2021/06/musicxml40/musicxml-reference/elements/harmony} (Retrieved March 26, 2025)} field into very long strings, which contradicts our motivations for the usage of the **kern format.
For these reasons, we developed an extension of the \emph{musicxml2hum} tool that converts the chords in a
representation that is suitable for our goals (see Section ~\ref{sec:chordformalizacion}).



\subsection{Musical score images}
We provide musical score images of two kinds: handwritten and synthetic. 

The handwritten scores were produced by jazz professionals and students at different levels, who were provided with printed versions of the digitised musical scores. They were instructed to copy the scores while maintaining the same chord symbols and layout (i.e., the same number of measures for each staff).
To simulate a realistic use case, participants were asked to either scan the completed scores or photograph them using a mobile phone.
All the collected scores were manually checked to assess their quality (more details in  Section~\ref{sec:quality}).
Some jazz standards were copied multiple times, for a total of 293 handwritten scores with an average length of 31.6 measures. To further align with real-world usage scenarios, we release the images in JPG format as they were received, retaining their varying resolutions, angles, and light conditions.

The synthetic part was produced with the Musescore 4 command line tool: for each piece in our digitised scores collection, we generated 2 synthetic full-page renderings: one with the Classical font and one with the MuseJazz font, which tries to mimic the human handwritten style that is commonly used for jazz music. The images were generated in SVG format, then the lyrics were removed, and finally images were converted to PNG.
In total, we have $326$ synthetic scores with an average length of 30.3 measures.

\subsection{From page to staff regions}

The data presented above can be used to train systems focusing on \textit{page}-level OMR. However, a large number of systems work at \textit{region} level, where each region consists of a music part that can be read sequentially, typically from left to right. 
For our lead sheets, a region would be a single staff with chord symbols above it.
This region-level approach simplifies the OMR task significantly, requiring smaller networks and less training data. Moreover, dedicated networks can be trained to segment the page into multiple regions, making this methodology both practical and effective when working with small datasets.

For this reason, we also include region annotations in the dataset, specifically bounding boxes that identify each staff with chords and a reference to their corresponding digitised staves. We created the bounding boxes using a region identifier system~\cite{yolov8_ultralytics} specifically trained for recognising music parts and manually checked them afterwards. 
The alignment of a region to the specific part of the **kern score has been obtained by splitting at the end of line token (\texttt{!!linebreak:original}), which is automatically inserted when converting from MusicXML using the above mentioned \emph{musicxml2hum} tool from Humlib and the \emph{yank} tool from Humdrum Tools~\footnote{https://github.com/humdrum-tools/humdrum-tools (Retrieved September 10, 2024)}. To keep the **kern source clean, and the region-wise prediction task reasonable, all comments and measure numbers have been removed.

Region-wise, our dataset consists of 2021 handwritten regions and 2208 synthetic regions with an average of 4.6 and 4.5 measures, respectively.

\subsection{Quality issues in the dataset}\label{sec:quality}

During the inspection of the handwritten scores, we identified the following quality problems.
Strike-through, hard-to-read calligraphy, and note-chord misalignments are frequent, but they are part of what makes this problem challenging, so we retain them. 

A large number of scores didn't respect the layout of the reference digitised score. This is problematic for page-level systems that aim at correctly transcribing the layout, and in particular for the extraction of region-level annotations.
To avoid discarding all this material, we decided to adapt the digitised scores, i.e., we created a different version of the MusicXML and **kern files that respect the layout of the handwritten score. For this reason, in our dataset, there are as many digitised scores as the number of handwritten scores, though the differences between multiple versions of the same digitised score may be minimal. Partially written scores were also treated in the same way.

Another kind of problem we identified is in the usage of equivalent chord symbols. In jazz lead sheets, there are multiple ways of writing the same chord, for example, a major 7 chord can be written as ``maj7'' or ``$\Delta$7'', or a minor chord can be written with ``-'', ``m'', ``mi'' or ``min''. Jazz musicians learn to read all these equivalent symbols and are not sensitive to the specific symbols used. Indeed, we found that, despite being instructed to use the same symbols, many transcribers used equivalent ones.
A solution to this problem would be to manually adapt all labels in the MusicXML scores to match the ones used by the transcriber. However, given the time-intensive nature of this task, we have prioritised other aspects of the dataset for this release, with plans to incorporate this in future versions.
An alternative path, which we follow in this paper, is to define the chord recognition problem as a mapping not between an image and a specific label, but between an image and the entire class of equivalent labels for a certain chord. In a use-case scenario, the choice of a specific label for the final output can be selected by the user depending on their preference.
We then need to identify sets of equivalent chords in a unique way, as we detail in the next section.

Finally, we address a quality issue in the chord encoding of MusicXML scores. We found that many scores used an encoding that did not follow the MusicXML standard. For example, instead of encoding a C6 chord with the harmony/kind attribute \texttt{major-sixth}, the harmony/text field was used with the ``6'' string. While the graphical output rendered from MusicXML is the same, this is an issue for the automatic processing of scores and our **kern converter. We manually corrected the MusicXML files to ensure a well-defined and consistent chord encoding.

\subsection{Chord formalization}
\label{sec:chordformalizacion}


To properly describe the chords in our **kern scores, we rely on the Harte syntax~\cite{harte2005symbolic}.
However, Harte syntax still allows multiple ways of encoding the same chord, for example, with combinations of different shorthands and extensions, or by specifying every single note. Therefore, we restrict the syntax as follows. Every chord is composed of a root, a shorthand, an optional list of extensions and an optional alternative bass, encoded in a string as:
\begin{equation}
    \textrm{root:shorthand(extension1,extension2,...)/bass}
\end{equation}
The list of shorthands we consider is 'aug', 'aug7', 'dim', 'dim7', 'hdim7', 'maj', 'maj11', 'maj13', 'maj6', 'maj7', 'maj9', 'min', 'min11', 'min13', 'min6', 'min7', 'min9', 'minmaj7', 'sus2', 'sus4', '11', '13', '7', '9'.
All remaining chords are obtained by adding an extension, which can add or remove a certain pitch, for example, C:7(b9). A chord for which a shorthand exists cannot be encoded with an extension; for example, C:maj(7,b9) is not a valid formulation in our syntax. The note removal in an extension, i.e., the subtract degree type in MusicXML\footnote{\href{https://www.w3.org/2021/06/musicxml40/musicxml-reference/data-types/degree-type-value}{https://www.w3.org/2021/06/musicxml40/musicxml-reference/data-types/degree-type-value} (Retrieved March 26, 2025)}, is coded with the ``no'' prefix, e.g., C:maj(no5,b9).



\subsection{Data split}
To facilitate the usage of this dataset for benchmarking, we release an official data split. We consider the number of unique pieces in the dataset, and assign 70\%, 10\%, and 20\% of them to the train, validation, and test splits, respectively. 
We ensure that all unique pieces for which multiple handwritten copies exist are placed in the train subset only. This ensures that we do not have data in different splits that share the same ground truth, which would bias the experiments, since the model could just memorise a piece and reproduce it.
Specifically, out of 163 unique pieces, we assign 115, 16, and 32 to the train, validation, and test subsets. When counting the number of handwritten scores (i.e., including multiple versions of the same piece), the three subsets include 245, 32, and 16 scores, and when counting the number of regions, 1696, 102, and 220.





\section{Model}\label{sec:methodology}
In this section, we describe our approach to OMR of jazz lead sheets. We focus on the region-level approach, i.e., our system's input is a single staff with chords on top.\footnote{Existing layout analysis methods can be successfully used to detect regions from full-page images \cite{dvorak2024staff}.}
Formally, each sample in our dataset consists of a pair $(\mathbf{x},\mathbf{y})$ of a region-level image
represented as matrix $x\in\mathbb{R}^{c \times h \times w}$,  and a sequence of musical symbols $\mathbf{y} = \left(y_{1},y_{2},\ldots,y_{|\mathbf{y}|}\right)$. Each sequence of symbols is drawn from a common vocabulary $\Sigma$ and is uniquely associated with a **kern file, through the tokenisation process described in Section~\ref{sec:tokenization}. 

While the score images are distributed with the original colors, we use black-and-white images, so we set $c=1$. We fix the dimension of each image to $h=128$ and $w=1000$ by first downsampling each image, maintaining the aspect ratio, and then right padding on the width dimension.

\subsection{Architecture}
We use the architecture proposed by Rios-Vila et al.~\cite{Rios24}
The architecture is an encoder-decoder model, where the encoder, a ConvNext~\cite{liu2022convnet} network, acts as a feature-learning block, by processing the image $\mathbf{x}$ into a compressed hidden representation $\mathbf{z}$. 
%
%
The decoder, is from Vaswani et al.\cite{vaswani2017}, and produces at each step $t$ a probability distribution over all symbols $\Sigma$, $p_t\in \mathbb{R}^{\left|\Sigma\right|}$, taking into account both the precedent tokens $(\hat{y}_0,\dots,\hat{y}_{t-1})$ and the image hidden representation $\mathbf{z}$:
\begin{equation}
\hat{p}_t = P(\hat{y}_t \mid z, (\hat{y}_0, \hat{y}_1, \hat{y}_2, \dots, \hat{y}_{t-1}))
\end{equation}

Prediction consists in concatenating the $\arg\max$ of $p_t$, i.e. $\hat{y}_t$, for all the steps. This sequence always starts with the special token $<bos>$ (begin of sequence) and the decoder stops predicting when $\hat{y}_{\left|\hat{y}\right|}=\;<eos>$ (end of sequence).

An advantage of using the same model as Rios-Vila et al. is that we can initialise our network with the weights from their publicly available pretrained checkpoint. Even though our tasks are not the same, they may still be related enough to make pretraining beneficial. The only part of the model we need to change is the last linear layer, since we use a different symbol vocabulary $\Sigma$.

\subsection{Tokenisation}\label{sec:tokenization}
\begin{figure}
  \centering
  \includegraphics[alt={Kern file with different tokenization strategies},width=0.85\linewidth]{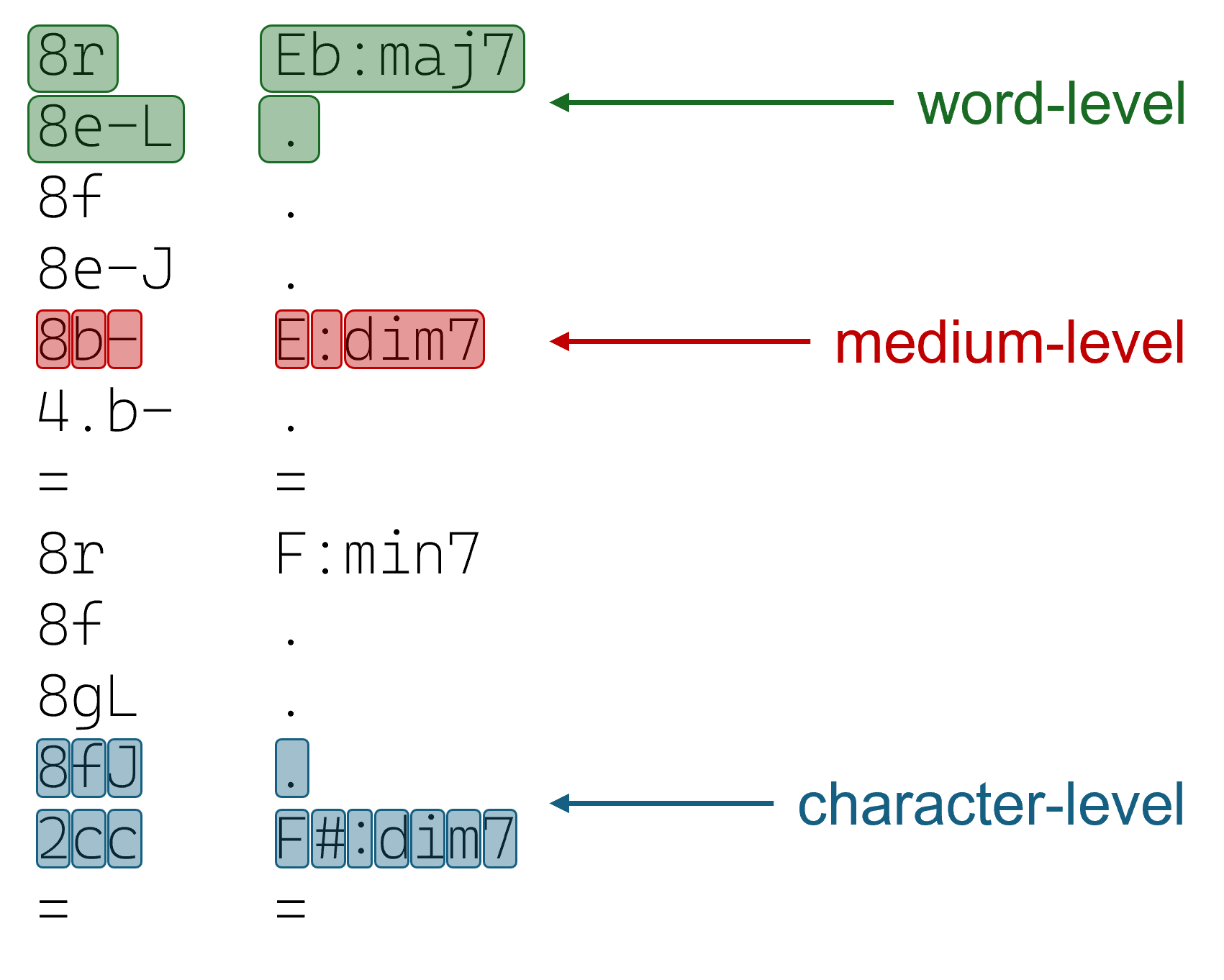}
  \caption{First two measures of \textit{Ain't Misbehavin'} in **kern format, with graphical examples of our tokenisation strategies.}
  \label{fig:tokenisation}
\end{figure}

As depicted on the left part of Figure~\ref{fig:tokenisation}, our kern scores are text files with two tab-separated spines, one for melody and one for chords. The melodic spine specifies the duration and pitch of the notes (or if there is a rest), and also other graphical elements, such as beaming starts (L) and ends (J).  There are multiple ways of using these files to train our neural network; we propose and evaluate three different tokenisation strategies for the **kern scores: \textit{word-level}, \textit{character-level}, and \textit{medium-level}, which build three different vocabularies $\Sigma_w$, $\Sigma_c$, and $\Sigma_m$ respectively.

The word-level tokeniser is the same employed by Rios et al.~\cite{Rios24}, and considers every tab or new-line separated string as a single token. This had the advantage of generating short sequences, at the cost of a very large (1762 tokens) vocabulary. With this tokenisation, the model is forced to learn independently how to process symbols, even when their graphical representation may be very similar; for example, a D\# and a D\# with a duration dot.

The character-level strategy treats every character as a single token. In direct contrast to the previous tokeniser, this yields long sequences and a small vocabulary (69 tokens). The system can more effectively reuse the graphical-to-symbol mapping it learns, for example, when a sharp symbol \# is used in a note or a chord. As a drawback, this forces the transformer decoder to pay more attention to the context, for example, the letter ``a'' corresponds to very different graphical symbols in the melody spine (a note) and chord spine (part of the ``maj'' or ``aug'' chord type).

We develop the last tokeniser to be on a medium level between the previous two. The goal is to have a bijective mapping between graphical symbols and tokens. For the note spine, we encode all pitches as unique symbols (even when they are made of multiple characters, for example, "ee" corresponds to E one octave higher than middle E), and all other characters as unique symbols, to separate flats, sharps, duration dots, ties, slurs, and beamings. For the chord spine, we separate the root, the type, the extensions (each extension is considered singularly if there are multiple), and the bass if they are specified. All other symbols, like time signature, key signature, clef, spine definition, and measure separators, are considered at the word-level. This generates a vocabulary of size 153. 

These tokenisers are developed to work in both directions, i.e., they produce tokens from a **kern file, to train the model, but they also reconstruct the **kern file from a model output for evaluation.

These tokenisers are straightforward to understand and can be implemented in just a few lines of code. This simplicity stems from the use of **kern files, which we use, instead of other formats, such as MusicXML, since they provide a compact and easy-to-handle representation of musical scores. 
However, the **kern format has its limits; for example, accidentals are explicitly encoded in each note, even if they are defined on the key signature, and this doesn't align with the graphical representation. For future work, one could have much more control over the tokenisation process by parsing the musical score into some dedicated internal representation, and then producing the tokens, similarly to what MidiTok~\cite{miditok2021} does with MIDI files.\footnote{Note that the tokens they produce are only focusing on the musical content, while we want to also retain graphical information.}
This would also allow us to work with different file formats which encode a wider set of graphical information.


\subsection{Metrics}\label{sec:metrics}
As evaluation metrics, we use edit-distance-based metrics, as used in other recent studies~\cite{martinez2023,Rios24}, which count the number of insertions, deletions, or modifications to transform our predicted **kern output into the ground truth. In particular, we consider 3 metrics: character error rate (CER), word error rate (WER)\footnote{Note that in the OMR literature WER is referred to as Symbol Error Rate (SER),
but we use WER to be consistent with the terms used in the tokenisation section.}, and line error rate (LER). 
The LER works at the line level, i.e., it counts an error if there is even a single-character difference between two lines. This is the only metric that allows us to evaluate if the chord is aligned to the correct note, but it doesn't consider ``almost-correct'' lines, for example, a line where a single duration dot is mispredicted. 
CER works at the character level, but it has the disadvantage of overly penalising wrong predictions that involve several characters, such as the chord type ``maj7''.
WER works at the word level, so it has similar disadvantages that LER, but less pronounced; for example, it will still consider correct a note, even if the chord next to it is mispredicted. All these metrics have different advantages and disadvantages, so we chose to report all of them. The task of finding a good unique metric that aligns with human perception is still an open problem in the field. 

\section{Experiments}\label{sec:experiments}


With the experiments in this section, we want to answer the following three research questions: Do pretrained SOTA weights on page-level polyphonic piano music~\cite{Rios24} improve the performance in our tasks? Which of the three tokenisers is more effective? Does the inclusion of synthetic data during training help the performance on real data?
All combinations of these three factors result in 12 experiments, which we describe in the following.



\subsection{Experimental settings}

We run our experiment for 100 epochs with a learning rate of $5\times{10}^{-4}$, with cosine annealing, a warmup phase of 150 steps, weight decay of 0.01, and batch size 64.
For each experiment, we use early stopping and compute our test results with the model that minimises the WER metric in the validation split. 
The model ConvNext encoder has three layers with kernel sizes of [3, 3, 9] and output channel sizes of [64, 128, 256]. The language model block, i.e., the decoder, is composed of 8 layers, each of which uses 4 attention heads and a hidden dimension of 256. 

\subsection{Results} \label{sec:results}

\begin{table*}[!ht]
\centering
\resizebox{.99\textwidth}{!}{%
    \begin{tabular}{lcccccccccccc}
    \toprule[1pt]
    \multirow{2}{*}{Metric} & \multicolumn{3}{c}{handwritten} & \multicolumn{3}{c}{handwritten + synthetic} & \multicolumn{3}{c}{pretrained + handwritten} & \multicolumn{3}{c}{pretr. + handw. + synth.} \\ 
    \cmidrule(lr){2-4} \cmidrule(lr){5-7} \cmidrule(lr){8-10} \cmidrule(lr){11-13}
    & word & medium & char & word & medium & char & word & medium & char & word & medium & char \\
    \cmidrule(lr){1-13}
    
    $\downarrow\,$WER    &    40.71     &     44.15     &    40.96    &    41.38     &     35.15     &    27.59    &    15.57     &     12.42     &  15.91  & 12.55  & \textbf{11.90} & 12.86\\

    $\downarrow\,$CER    &   52.93      &   55.51       &    52.52    &    53.67     &     45.70     &   34.78     &    19.43     &     14.35     &    18.40   & 16.37 & \textbf{13.67} & 15.32\\

    $\downarrow\,$LER    &     75.01    &     79.40     &   79.54     &    73.55     &    71.60      &    57.84    &     39.94   &     31.28     &     39.49  & 32.99 & \textbf{29.68} & 32.71\\ 
    \bottomrule[1pt]
    \end{tabular}
}
\caption{Results of our experiments with different tokenisers (at word, medium, and character level), the addition of synthetic data, and the usage of a pretrained model.}
\label{tab:results1}
\end{table*}

    



The results of our experiments are summarised in Table~\ref{tab:results1}
Initialising our model with the pretrained checkpoints before training brings clear advantages for every metric, tokeniser, and data setting. Without pretraining, our model was not able to properly train with handwritten data only, and stayed at a WER level $>40$. We found that this corresponds to a model that has learned the basic **kern syntax (i.e., it places the notes and chords properly to the left and right spine, respectively), but doesn't seem to consider the input image at all. Adding synthetic data helped to ease this problem, but only for the medium and character-level tokenisers, and the performance was still much worse.

Regarding the choice of the tokenisers, the medium-level one yields the best results across all metrics for the pretrained models. Similarly, using synthetic data is also beneficial. The results are less clear for the non-pretrained models, but as said before, these models did not properly train, so we should not trust a comparison between their outputs, since we may as well be analysing just some random noise.
According to all three metrics, our best model is the model trained from the pretrained checkpoint, with handwritten and synthetic data and using the medium-level tokeniser.

We want to emphasise that while our results are clear and consistent across our experiments, they are only valid for our dataset and model choices. We can speculate that as the model and dataset get bigger, the advantage of using a well-balanced tokenisation procedure would gradually diminish, in favour of flexible tokenisation like the character-level one, which can handle all kinds of **kern scores; or even be able to directly produce very verbose file formats like MusicXML. However, if one is interested in efficient systems or needs to work in a low-data scenario, it is worth exploring these music-informed directions.

\begin{figure}[t]
    \centering
    \includegraphics[alt={Image of a Embraceable You prediction.},width=0.85\linewidth]{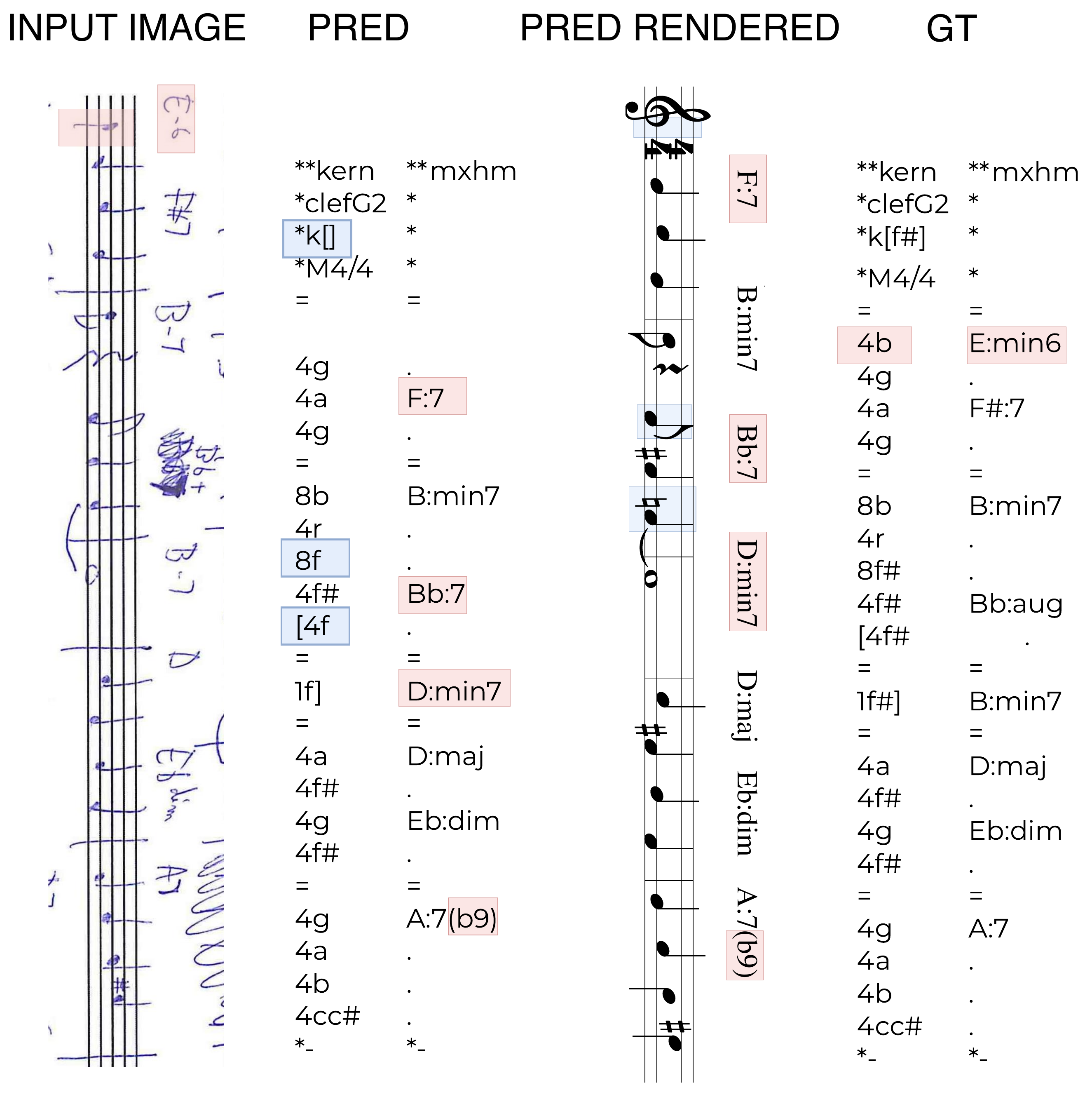}
    \caption{Prediction of \textit{Embraceable You} by George Gershwin (bars 10th to 14th) excerpt with our best model. In red, errors associated with transcription or missing symbols. Blue boxes depict semantic errors.}
    \label{fig:comparison}
\end{figure}

\subsection{Qualitative analysis}

In Figure \ref{fig:comparison}, we show an example of the transcription performance of our best model (pretrained, with synthetic data and medium-level tokenisation).

Attending to red boxes, which represent both missing and wrongly transcribed symbols, it is worth highlighting that most of the errors we encounter are related to chord symbols. This behaviour outlines the difficulty of transcribing simultaneously both sources (melody and chords). 
We can speculate that this derives from the higher frequency of notes compared to chords in a jazz lead sheet, so the model has much less chance to learn a correct behaviour for chords during training.
Moreover, the use of the pretrained model on the polyphonic grand piano dataset biases the framework towards transcribing the melody, as no chords appear in that dataset. On the positive side, the chord-note alignment seems to work pretty well, even in the case of the strike-through in measure 2.

Finally, we highlight in blue the semantic errors. These are common problems in OMR transcriptions, where some symbols are transcribed correctly if we only consider their local graphical features, such as the note ``1f'' at the start of the third bar. Graphically, it does not contain the sharp symbol (\#); however, given the key of the piece (\textit{*k[f\#]}), the model misses this implicit accidental.

\section{Conclusion}\label{sec:conclusion}

With this paper, we took an initial yet significant step towards the development of OMR systems for handwritten jazz lead sheets that could benefit musicians and MIR researchers.
We collected a dataset of 293 handwritten lead sheets produced by music professionals and students of different levels, addressed its quality problems, aligned it with ground truth digitised scores in MusicXML and **kern formats, and organised it in a suitable format for training end-to-end page-level and region-level systems. We also supplied synthetic scores generated from the ground truth and a data split to facilitate the usage of this dataset for benchmarking.
We released the first OMR system dedicated to handwritten jazz lead sheets, based on an encoder-decoder architecture, which is used for polyphonic piano OMR. We proposed multiple tokenisation strategies and analysed their effect, showing that the most effective option (at least for our model and dataset size) is to have a unique mapping between graphical symbols and tokens, without multiple symbols being grouped in a single token, or a single symbol being encoded by multiple tokens. We also proved that the usage of synthetic data is beneficial, and that pretraining our network with the different but related task of polyphonic piano transcription is fundamental to enable effective training on our dataset.

Since the performance of deep-learning systems highly correlates with the size of the training data, future work on the dataset part would include expanding the collection of handwritten scores and including new digitised scores to produce more synthetic data. Since we are also distributing the SVG files for the synthetic scores, multiple data augmentation procedures can be performed, such as changing the colour and the width of lines, and applying different backgrounds and image transformations to simulate real paper in different lighting conditions.
On the model side, it is reasonable that employing more recent transformer components (e.g., rotary positional embedding) could make our model more performant and less data-hungry. The usage of a vision transformer instead of the ConvNext encoder would also be a natural next step in the scaling of our network. Finally, as we found it highly beneficial to use a pretrained checkpoint on another OMR task to kickstart our training, we believe that incorporating a wider range of OMR tasks in the pretraining phase could further enhance the results. This also motivates the development of a general OMR model capable of functioning across various music domains.

\section{Acknowledgments}
We thank `Centro Superior de Música del País Vasco, Musikene' (Itziar Larrinaga), `Casa Sofía de El Altet, Alicante' (Pedro J. Ponce de León), `Conservatorio Superior de Música Joaquín Rodrigo de Valencia' (Jorge Sevilla), `Sedajazz, Valencia` (Francisco A. Blanco Latino),  `Conservatorio Superior de Música Óscar Esplá de Alicante' (Manuel Mas), and `AMCE Santa Cecilia' for coordinating their students’ meticulous work in transcribing jazz lead sheets. Patricia Garcia-Iasci holds a research contract with the University of Salamanca funded by the Regional Government of Castilla y León (Order EDU/1009/2024 of October 10) and co-financed by the European Social Fund Plus (FSE+) budget code (18.218F 463AB05). This paper is supported by grant CISEJI/2023/9 from ``Programa para el apoyo a personas investigadoras con talento (Plan GenT) de la Generalitat Valenciana'', the European Research Council (ERC) under the EU's Horizon 2020 research \& innovation programme, grant agreement No.\ 101019375 (\textit{Whither Music?}), and the Federal State of Upper Austria (LIT AI Lab).




\bibliography{ISMIRtemplate}

\end{document}